\documentclass[conference]{IEEEtran}
\IEEEoverridecommandlockouts
\usepackage[caption=false,font=footnotesize]{subfig}
\usepackage[rightcaption]{sidecap}
\usepackage{amsmath,amssymb,amsfonts,bm}
\usepackage{algorithm,algpseudocode}
\usepackage{graphicx}
\usepackage{url}
\usepackage{flushend}

\usepackage[square,numbers]{natbib}
\usepackage{xcolor}
\def\BibTeX{{\rm B\kern-.05em{\sc i\kern-.025em b}\kern-.08em
    T\kern-.1667em\lower.7ex\hbox{E}\kern-.125emX}}
\begin{document}

\title{ Bias Remediation in Driver Drowsiness Detection systems using Generative Adversarial Networks
}

\author{Mkhuseli Ngxande, Jules-Raymond Tapamo, \IEEEmembership{Member, IEEE}, and Michael Burke, \IEEEmembership{Member, IEEE}}
\maketitle

\begin{abstract}
Datasets are crucial when training a deep neural network. When datasets are unrepresentative, trained models are prone to bias because they are unable to generalise to real world settings. This is particularly problematic for models trained in specific cultural contexts, which may not represent a wide range of races, and thus fail to generalise. This is a particular challenge for Driver drowsiness detection, where many publicly available datasets are unrepresentative as they cover only certain ethnicity groups. Traditional augmentation methods are unable to improve a model’s performance when tested on other groups with different facial attributes, and it is often challenging to build new, more representative datasets. In this paper, we introduce a novel framework that boosts the performance of detection of drowsiness for different ethnicity groups. Our framework improves Convolutional Neural Network (CNN) trained for prediction by using Generative Adversarial networks (GAN) for targeted data augmentation based on a population bias visualisation strategy that groups faces with similar facial attributes and highlights where the model is failing. A sampling method selects faces where the model is not performing well, which are used to fine-tune the CNN. Experiments show the efficacy of our approach in improving driver drowsiness detection for under represented ethnicity groups. Here, models trained on publicly available datasets are compared with a model trained using the proposed data augmentation strategy. Although developed in the context of driver drowsiness detection, the proposed framework is not limited to the driver drowsiness detection task, but can be applied to other applications. 
\end{abstract}


\section{Introduction}
The ability of Artificial Intelligence systems (AI) to automate decision-making capabilities in human daily lives is increasing rapidly. As a result, these systems influence human interaction with the real world and are transforming the future. Their decision-making capabilities typically rely on large training datasets that learn and extract useful patterns in an automated way. Unfortunately, if these systems are trained on datasets that do not have a complete representation of real-world scenarios, they may be prone to bias and prejudice society. 

One application where the use of deep learning techniques is increasingly gaining popularity and in which unrepresentative training datasets can lead to negative consequences is that of driver drowsiness detection. One of the primary objectives of the motor industry is passenger safety. Road related accidents are a primary cause of injuries and death among the human population \cite{globalreport}. Among the factors leading to accidents, driving while drowsy is of particular concern. This has prompted the automobile industry to make efforts to develop detection systems that improve driver safety. Monitoring driver behaviour through computer vision and machine learning techniques that detect drowsiness and warn the driver is an increasingly popular technique under investigation by the motor industry. Statistics reveal that a high rate of accidents is caused by drowsy drivers and 20\% of serious accidents arise from a failure of driver's judgement and their inability to control the vehicle in the drowsy state \cite{driver2019}. In addition, the World Health Organisation (WHO) reveals that deaths arising from road traffic crashes have increased to 1,35 million in the year of 2018 \cite{WHO2019}. The report by WHO further shows that nearly 3 700 people die on roads every day. This is a particular concern in Africa, which only has 2\% of the world's cars, but has the highest accident rates of about 20\% of road deaths \cite{Africa2019}. These are alarming findings, which urgently need to be addressed. However, the development of robust driver drowsiness detection systems is still a challenge in both academia and industry. \par

In the automobile industry, several attempts have been made to monitor driver's state over time by considering various methods such as vehicle-based measures, physiological signals, and behavioural measures. Vehicle-based methods make use of car electronics together with appropriate sensors \cite{Knipling1994}. These sensors are usually placed on the pedals, steering wheels and often include cameras around the car \cite{Sahayadhas2012}. Unfortunately, these methods mostly rely on the state of the car and the surrounding environment and focus less on the driver's state. On the other hand, physiological methods monitor the driver's state using physiological signals which can be recorded using devices such as Electroencephalogram (EEG), Electrooculogram (EOG), Electrocardiogram (ECG), or Electromyogram (EMG) \cite{Luthra2015,Drewes2000,Folane2016}. These devices yield accurate results because the readings measure brain activity \cite{Awais}. However, physiological methods are invasive as they typically require a device to be placed on the driver to record signals. These devices can make the driver uncomfortable and are considered impractical for real-time drowsiness detection systems. \par
In contrast, behavioural methods are non-invasive methods that make use of a mounted camera to track the face of the driver and measure the level of drowsiness based on facial features. There are numerous facial features that can be used to measure drowsiness from the camera feed including eye state, yawning, and head position \cite{Man2015, Rajput2013, JACOBEDENAUROIS2017}.  Behavioural methods can be combined with machine learning techniques to produce more robust systems. A meta-review \cite{8261140} examined Hidden Markov Models (HMM), CNNs, and Support Vector Machines (SVM) and concluded that CNNs performed better than other techniques, although SVMs were most commonly used. The success of CNNs has been shown in many computer vision tasks including object classification, segmentation, and object tracking \cite{8279704, Jangid2018,7978672}. CNN architectures require a large amount of training data to learn a suitable representation for a given task. Unfortunately, when it comes to driver drowsiness detection, there are a limited number of publicly available training datasets, and some datasets are not published because of security and privacy reasons preventing the publication of people's faces. In addition, publicly available datasets are often unrepresentative as these may not cover a wide variety of ethnicities. For African contexts, this poses a challenge since the population is diverse and individuals can have many different facial attributes. Models trained using publicly available datasets do not generalise well in an African context \cite{detecting-inter}. \par

The limitations of datasets that fail to cover a wide range of ethnicities lead to bias in trained models when it comes to contexts with different nationalities. Prior work has shown that visualisation techniques can be used to identify bias in training datasets by identifying population groups where a classifier tends to fail \cite{detecting-inter}. This paper makes the following contributions in addressing population bias in driver drowsiness training datasets:

\begin{itemize}
\item Introduces a novel framework that remedies generalisation failures in under represented population groups in the training dataset, which boosts the performance of drowsiness detection across all population groups.
\item Introduces a sampling algorithm that identifies individuals with facial features where the network is failing.
\item Shows how a GAN that generates realistic images can be used to produce training data for those races or individuals where the model is failing.
\end{itemize}
 The framework relies on two primary components, population bias visualisation and a Generative Adversarial Network (GAN). The GAN generates realistic images of individuals (drowsy and awake) in these population groups, which are used for retraining the ResNet model used for drowsiness detection with new parameters i.e learning rate and epoch sizes to reduce overfitting and improve the detection accuracy. The population bias visualisation is used to group races by similarity and identifies where the model is failing to generalise. This process is iteratively repeated until convergence.\par
This paper is organised as follows. Section \ref{BACKGROUND} provides an overview and background work, which is followed by a discussion on GANs, population bias and CNN visualisation. Details around our framework are discussed in section \ref{Implementation}, and with training information and  a description of the datasets used in this paper. Experimental details and results are presented in section \ref{EXPERIMENTAL}. Finally, Section \ref{CONCLUSION} provides a brief conclusion of our work.

\section{BACKGROUND AND RELATED WORK}\label{BACKGROUND}
 
 In this paper, a novel framework that boosts the performance of CNNs for driver drowsiness detection is presented. This is accomplished by highlighting regions where the model is failing and passing similar GAN generated images to the model for retraining. This strategy is based on boosting,  where a weak classifier is iteratively re-weighted to make it a strong classifier. In this section, we explain related works on data augmentation and visualisation strategies.
 
\subsection{Generative Adversarial Networks}

Since their introduction in 2014 by Goodfellow et al. \cite{goodfellow2014generative}, GANs have shown great success in many computer vision tasks, including pose guided person image generation \cite{IsolaZZE16}, domain transfer \cite{1811-07056}, super-resolution \cite{00219}, and text to image applications \cite{00676}. Many variants of GAN architectures have been developed, such as Wasserstein Generative Adversarial Networks (WGAN) \cite{Arjovsky2017}, Wasserstein Generative Adversarial Networks with Gradient Penalty (WGAN-GP) \cite{GulrajaniAADC17}, and Deep Convolutional Generative Adversarial Network (DCGAN) \cite{RadfordMC15}. These architectures have been proposed to improve the original GAN architecture for various tasks. \par
The original GAN architecture is composed of two neural networks namely, a generator $G$ and a discriminator $D$ which are trained by playing a mini-max game against one another. In the case of data augmentation, we utilize the domain shift of one image to another domain. In the transfer of awake states to drowsy states, where we seek to learn a generator distribution $\textit{p}_{g}$ over data $x$, the generator creates a mapping function, parameterized by $\theta_g$ from a prior latent distribution $\textit{p}_{z}(z)$ to data space $G(\textit{z;}\theta_{g})$. The discriminator $D(x;\theta_{d})$, on the other hand, learns parameters $\theta_d$ to distinguish whether images are from the training data or from the generator. The mini-max game function  $V(G,D)$ is expressed as follows:
\begin{IEEEeqnarray}{lCl}
min_{G} \hspace{1mm} max_{D} \hspace{1mm} V(D,G)=\mathbb{E}_{D} + \mathbb{E}_{G} \\
\text{where } \mathbb{E}_{D}= \mathbb{E}_{\textit{x}\backsim \textit{p}_{\textbf{data}}(\textit{x}) }[ \log D(x)] \nonumber\\
\mathbb{E}_{G}= \mathbb{E}_{\textit{z}\backsim \textit{p}_{\textit{z}}(\textit{z})}[\log(1-D(G(\textit{z})))] \nonumber
\label{mini}
\end{IEEEeqnarray}

 Unfortunately, the original architecture is limited in that there is no flexibility in generating desired outputs. To overcome this problem, conditional GANs were introduced as an extension that introduces additional information to both the networks \cite{mirza2014conditional}. This additional information allows the flexibility of producing controllable outputs from the training dataset. The additional information, $y$, is typically a label which is applied to the resulting image, for example in our case this is the awake or sleepy state. The mini-max objective function from equation (1) is then updated as follows: 
 \begin{IEEEeqnarray}{lCl}
 min_{G}max_{D} V(D,G)=\mathbb{E}_{\textit{{x}}\backsim \textit{p}_{\textbf{data}}(\textit{{x}}) }[\log D(\textit{x$\vert$y})] + \nonumber\\ \mathbb{E}_{\textit{z}\backsim \textit{p}_{\textit{z}}}(\textit{z)}[\log(1-D(G(\textit{z$\vert$y})))]  \label{conditionnal}
\end{IEEEeqnarray}

In this paper, a controllable GAN is used as a data augmentation technique to balance the training dataset for a driver drowsiness detection task. Data augmentation is a technique that increases the size of a training dataset to reduce the chances of overfitting by the network. In computer vision, the most common way to perform data augmentation is by applying parameterised transformations including random cropping, rotation, scaling, and jittering. In the case of driver drowsiness detection, many available datasets are unrepresentative as these are often captured in specific cultural contexts. In addition, there is a distinct lack of datasets captured in African contexts \cite{6117593}. Applying common data augmentation strategies to the training dataset improves the results by a small amount. However, standard augmentation is severely limited and is unable to generalise to more complex domain shift problems. 

There is much work using GAN architectures for data augmentation. Gupta \cite{Gupta2019} used conditional GANs for sentiment classification, obtaining a significant improvement against a baseline model that was only trained on real data. The conditional GAN was trained using different strategies including pre-training and noise injection on the training data. Mok and Chung \cite{s11291} proposed an automatic data augmentation that enables machine learning methods to learn from the available annotated samples efficiently. Their architecture consists of a coarse-to-fine generator which captures the manifold of the training sets. Their proposed method was used on Magnetic Resonance Imaging (MRI) images and achieved improvements of about 3.5\% over the traditional augmentation approaches that were compared against. In addition, Wu et al. \cite{Wu} used a multi-scale class conditional GAN to perform contextual in-filling to synthesize lesions onto healthy screening mammograms. For experimentation, three classifications were compared and their method substantially outperformed a baseline model. Antoniou et al. \cite{Antoniou} used a conditional GAN to augment data in another domain. They named their architecture Data Augmentation Generative Adversarial Networks (DAGAN), which can be trained for low-data tasks using standard stochastic gradient descent approaches. It is clear that GANs can be used as a substitute for traditional augmentation techniques and are particularly valuable where more sophisticated augmentation strategies are required.

\subsection{Population Bias}

Racial bias is a problem that has been raised in the computer vision community, with a specific focus on how to develop machine learning models that guarantee fairness in all ethnicity groups. This problem has been identified as a result of investigations of fairness in machine learning systems that involves humans. Racial bias has been reported in various areas including criminal justice, employment, education, and face recognition systems \cite{Shadowen2017,Rhue2019,Rekognition2019}. This bias comes from unbalanced training datasets that favor the demographics of the contexts that the application were developed for. As a result, when tested outside of these conditions these algorithms begin to fail. In addition, publicly available datasets often do not capture a wide range of races, while policies of publishing data which contains people's faces prevents some datasets to be published. 

Buolamwini et al. \cite{buolamwini18a} found that the performance of three commercial gender classification algorithms decreases dramatically for darker-skinned female faces. Moreover, in \cite{Garvie2016} it is revealed that Amazon's facial recognition tool misidentified photos of 28 US parliamentary members as criminals because of their skin complexion. De-Arteaga et al. presented a large scale study of gender bias in occupation classification \cite{DBLP1901-09451}. Here, a machine learning algorithm learns to classify gender based on first names and pronouns from online biographies. Their study showed that there are gaps when using three different semantic representations such as bag-of-words, Deep Recurrent Neural Networks (DRNN), and word embedding \cite{DBLP1901-09451}. Benthall and Haynes have investigated supervised learning algorithms and revealed that they are exposed to racial bias because of the differentiation that is embedded in systematic patterns \cite{Benthall3287575}. 

 Abiteboul \cite{Abiteboul:2017} investigated issues in ethical data management, where he considered bias and violation of data privacy in data analysis. His work discusses factors to be considered when working with data such as fairness, transparency, neutrality, and diversity. To overcome the challenges this introduces, he has proposed a unsupervised learning technique that dynamically detects patterns of segregation in order to mitigate the root causes of social disparities and other factors that can lead to biased models.
 
In this paper, we seek to address the racial bias introduced by imbalanced training sets, by generating images of awake and drowsy people that look like those for whom the model is failing, so as to improve generalisation.

\section{Implementation}\label{Implementation}

\begin{figure*}
\centering
\includegraphics [width=\textwidth]{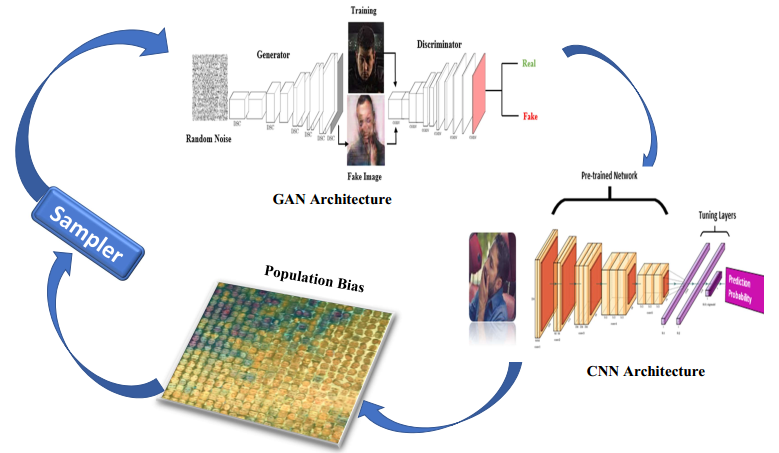}
\caption{The figure shows the proposed bias remediation framework. The first step uses a GAN architecture to generate synthetic data, which is used to train a CNN to detect driver state. A population bias visualisation is applied to the testing dataset and highlights where the model is failing to generalise. Images are then sampled where the model is not generalising, and are used to find similar images from the GAN generated images to continue training the model. \label{framework}}
\end{figure*}

This section discusses our framework. Figure \ref{framework} illustrates the general architecture of the framework. A pre-trained Resnet model is used for classifying the driver's state, after fine-tuning the final layers with fully connected layers and binary classification layer. The framework boosts the performance of the Resnet classification model on population groups where it is failing (in our case this is darker skinned individuals). 

The framework is composed of four primary components. Firstly, a GAN architecture produces synthetic images of individuals with facial attributes that can be used when retraining the network. The second component is the CNN architecture that predicts the state of the driver, while the third component is a population bias visualiser that highlights regions where the model is performing well and where it is failing to generalise. Lastly, the sampler targets images where the model is not performing well and searches through the synthetic images to find more similar images to those, which are used for model retraining. 

\textbf{\normalsize GAN architecture -} We adopt the architecture for our generative network from Choi et al \cite{star09020}, who have shown impressive results for generating realistic synthetic images in different domains. Our architecture is a conditional GAN that is conditioned to translate facial attributes such as awake or drowsy across multiple ethnicity groups. This translation of attributes helps in improving the detection model by supplying images with appropriate features where the drowsiness detector fails to generalise. The generator was modified by replacing the standard convolutions by depthwise separable convolutions \cite{DepthwiseGANs}. The benefit of this is to have fewer trainable parameters, while retaining the performance of the network. Furthermore, the generator consists of a stride size of two for down-sampling and 11 depthwise separable convolutions. We used instance normalisation \cite{ulyanov1607instance} instead of batch normalisation for the generator. For the discriminator network, standard convolutions were retained because the discriminator acts as a classifier and requires greater capacity in order to distinguish between real and fake images. In addition, PatchGANs \cite{Li2016} were adapted for discriminator network because they make use of a fixed-size  patch  discriminator that is  easily  applied to 256x256 images.

\textbf{\normalsize ResNet architecture -} We used a pre-trained ResNet model which comprises 50 layers and was originally trained on Canadian Institute For Advanced Research (CIFAR-10), ImageNet Large Scale Visual Recognition Competition (ILSVRC) and  Common Objects in Context (COCO 2) \cite{2015arXiv151203385H} datasets. We fine-tuned the pre-trained model on the last layers, but modified the prediction function to a sigmoid for binary drowsiness classification. 

\begin{figure}
\centering
\includegraphics [width=0.49\textwidth,height=6cm]{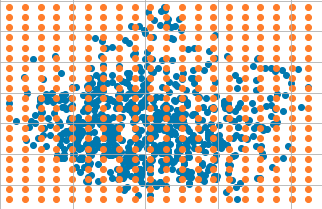}
\caption{The figure shows 2-dimensional grid projection coordinates obtained after applying the linear PCA transformation into 2-dimensional subspace. A uniformly spaced grid is placed over the projected image coordinates, and images are assigned a grid position by finding the closest image coordinate to each grid position, ensuring each image can only be used once. The blue dots represent grid position and the red dots represent PCA projections. \label{grid}}
\end{figure}

\textbf{\normalsize Population bias visualisation -} The population bias visualisation component relies on PCA \cite{Smith2002} for dimension reduction. This is followed by sorting the images by similarity and overlaying the prediction error to visualise where the model is failing. Image features are extracted by projecting validation images onto a 2-dimensional grid using PCA. The validation dataset is transformed into an orthogonal subspace where axes (Principal Components) align with the directions of maximum variance in the data. Here, a matrix of images is formed by reshaping images $\textbf{x}_{i}$ into row vectors (where $ \textit{i} = 1 ... \textit{N}$, and \textit{N} is the number of images in the dataset) and stacking these vertically to form an \textit{N} x \textit{P} matrix. The number of pixels in each image is denoted by \textit{P}. The PCA data transformation starts by mean centering the matrix of images, which is accomplished by subtracting the mean image from each 1 x \textit{P}  dimensional row vector, $\textbf{X}_{i}$ in the matrix of images,\begin{IEEEeqnarray}{lCl}
 \mathbf{\hat{X}}_{i} = \textbf{X}_{i}-\mu_{i}  
\end{IEEEeqnarray}
where $ \mu = (\mu_{i},..,\mu_{p})$ and $ \mu_{i}= \dfrac{1}{N} \sum ^{N}_{i}\textit{\textbf{X}}_{ij}$ and $ X_{ij}$ is the $j^{th}$ pixel of image.
The mean centred matrix  \textit{N} x \textit{P} dimensional matrix of images $\hat{\textbf{X}} $ is then decomposed using singular value decomposition (SVD),\begin{IEEEeqnarray}{lCl}
 \mathbf{\hat{X}} = \mathbf{U}\mathbf{\Sigma}\mathbf{V}^\text{T}
\end{IEEEeqnarray} 
Here, $\mathbf{U}$ is a $ P \times N$ unitary matrix, $\mathbf{V}$ is a $P \times P$ unitary matrix  and $\mathbf{\Sigma}$ is a diagonal matrix comprising the singular values of $\mathbf{\hat{X}}$ in decreasing order \cite{Stewart}. A reduced dimensional representation of $\mathbf{\hat{X}}$ can be obtained by discarding columns of $\mathbf{U}$ and $\mathbf{V}$,\begin{IEEEeqnarray}{lCl}
 \mathbf{\hat{X}} \approx \mathbf{U}_{0:j}\mathbf{\Sigma}_{0:j,0:j}\mathbf{V}_{0:j}^\text{T}
\end{IEEEeqnarray}
Here, $j$ denotes the number of columns retained. As shown above, PCA can project data into a low dimensional coordinate system, with axes provided by the columns of $\mathbf{U}_{0:j}$, and data coordinates given by $\mathbf{V}_{0:j}$.
In this work, we retain only two columns ($j=2$), and project images into a two-dimensional coordinate system. Figure \ref{grid} shows the 2D projection (coordinates obtained from $\mathbf{V}_{0:2}$) of facial images in our test dataset. We use this projection to construct a grid of images, grouped by similarity.\par
\begin{figure}
\centering
\includegraphics[width=0.49\textwidth,height=9cm]{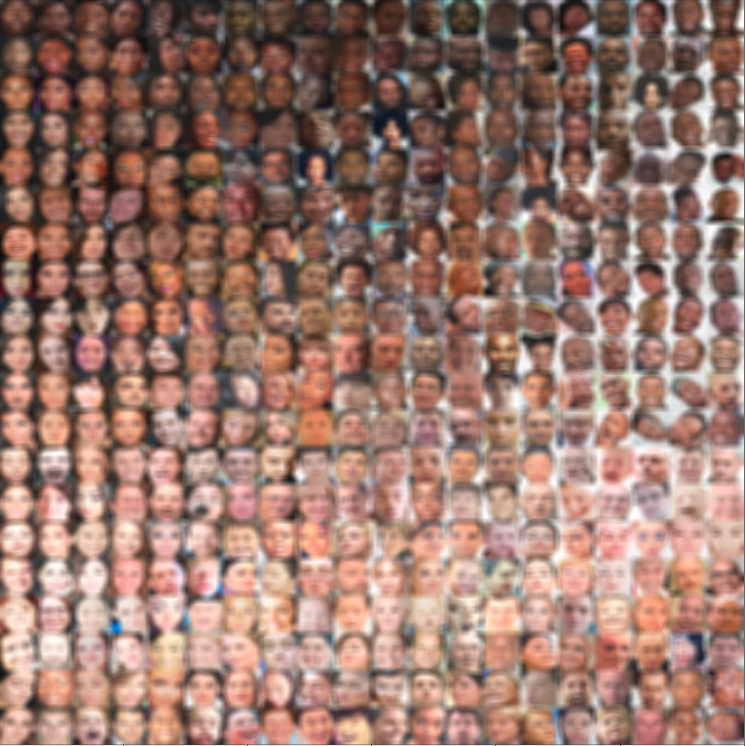}
\caption{The population bias visualisation strategy applies PCA to sort faces by similarity without requiring meta-data. \label{PCA_results}}
\end{figure}

  We create a uniform coordinate grid and search for the closest image (in the reduced dimensional coordinate system) to each point in the grid. We assign each image a corresponding point, and ensure that no image is duplicated, by removing it from the list of available images once allocated a grid coordinate in order to produce a grid of images that groups individuals by facial similarity, as shown in Figure \ref{PCA_results}. It is clear that this process successfully groups faces of similar state and complexion together, with darker skinned individuals located towards the top of the image, and lighter skinned individuals towards the bottom. For each image selected, we calculate the error in prediction, to produce a saliency map indicating model quality for the constructed grid of images.


\textbf{\normalsize Image Re-sampling -} This step is performed by randomly selecting images according to the error in prediction, termed failure probability here. The failure probability 
\begin{equation}
C_i = |y_i-y_t|
\end{equation}
is calculated by determining the difference between the CNN sigmoid prediction output, $y_i$ and the true label $y_t$ (a binary label encoding drowsy or awake).

This is normalised by the total probability of failure over all $N$ images in the dataset
\begin{equation}
\hat{C}_i = \frac{(1-C_i)}{\sum_{i=1}^N(1-C_i)}
\end{equation}

The random selection is performed by categorically sampling images using the weights above. This ensures that images with greater probability of failure are more likely to be sampled. Similar images in the GAN generated dataset are then selected by finding close matching images in the PCA space. The selected images are then used to continue training the ResNet model to increase classification performance. 

\subsection{Training}
In order to train the GAN architecture, the Adam optimiser \cite{2014arXiv1412.6980K} was used with $\beta_1 = 0.3$ and $\beta_2 = 0.6$, along with a batch size of 32. We applied a horizontal flipping data augmentation with a probability of 0.5. To train the model, we used a learning rate of 0.0001 for 100 epochs and then linearly decayed the learning rate over the next 100 epochs. This strategy compensates for the fact that our training data is limited. All experiments were carried out on a single Nvidia Tesla K20c GPU, where the training took approximately 12 hours.

Initially, the models were trained on three different datasets (NTHU-drowsy, DROZY, and CEW) and were compared with our framework as illustrated in the results section. The pre-trained ResNet model initially used a learning rate of 0.0001, which was then modified for the rest of the experiments from ${\rm 1e}^{-3}$ to ${\rm 1e}^{-6}$, which was performed for each iteration on our framework. We also used an early stopping strategy to prevent overfitting of the model.

\subsection{DATASETS}

This section describes the datasets that were used for training and testing ResNet model. For this work, the NTHU-drowsy, DROZY, and CEW datasets are used.

\begin{LaTeXdescription}
\item[NTHU-drowsy] was introduced at the 13th Asian Conference on Computer Vision (ACCV2016) \cite{NTHU}. The dataset is split into test and training sets. For training, there are 18 participants (10 men and 8 women) pretending to drive, with 5 scene scenarios for each participant including no-glasses, glasses, glasses at night, no glasses at night, and sunglasses. For evaluation, there are images of 2 men and 2 women. Videos combining drowsy, normal and sleepy states are provided.

\item[DROZY] consists of 14 participants (3 males and 11 females) \cite{massoz2016ulg}. Each video is approximately 10 minutes long and is accompanied by the results of psychomotor vigilance tests (PVTs) regarding the drowsiness state.  For each participant, the dataset contains a time-synchronized Karolinska Sleepiness Scale (KSS) score \cite{massoz2016ulg}.

\item[CEW] is a collection of online images of different races (for example Asians and non-Asians with light-skinned faces) and contains about 2423 participants \cite{ClosedEyes}. Among the participants, 1192 have both eyes closed and 1231 have their eyes open. These images were selected from the labeled faces in the wild database.
\item[Validation Dataset] Our validation set contains 1500 faces which some are obtained on the validation sets of the three datasets used. To have a balance and representative validation dataset, we added African faces which we collected for this purpose of drowsiness detection task. These images contain many ethnicity groups with facial drowsiness states.
\end{LaTeXdescription}

\section{EXPERIMENTAL RESULTS}\label{EXPERIMENTAL}
We first evaluated our proposed framework to results on the publicly available dataset using the pre-trained ResNet models. All the parameters were kept the same for all the first experiments, but thereafter the learning rate and training epochs were modified to prevent overfitting of the models.

\subsection{GAN Augmentation Results}
\begin{figure}
\centering
\includegraphics[width=0.40\textwidth]{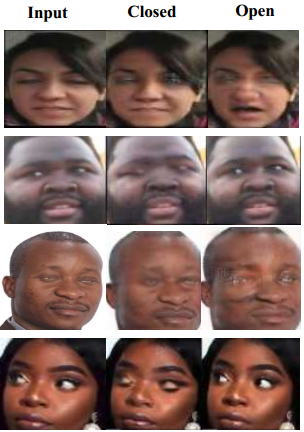}
\caption{Despite being imperfect, examples generated by the GAN serve to improve model performance.  \label{Complexion}}
\end{figure}

Figure \ref{Complexion} shows the eye state attribute transfer results on validation dataset. We observed that it is easier to transfer from eyes open to eyes closed. The first row in figure \ref{Complexion} shows that the network was not performing well in transferring from eyes closed to eyes opened, as shown by the blurriness of the image.  
Despite these limitations, these images still boosted the performance of detecting driver drowsiness task.

\begin{figure*}
\centering
\subfloat[ResNet-CEW]{\includegraphics[width=0.34\textwidth,height=6cm]{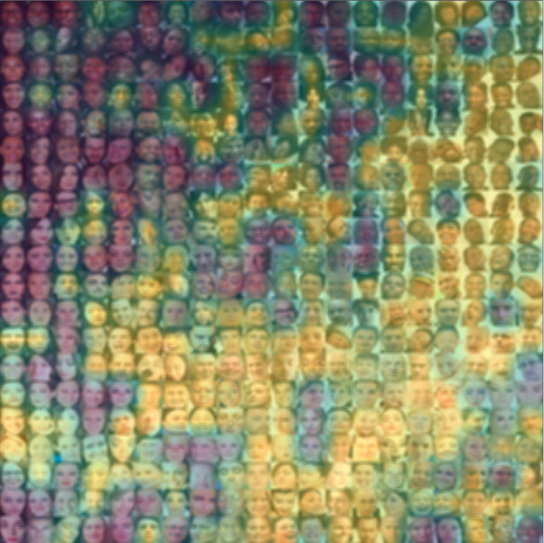}}
\subfloat[ResNet-DROZY]{\includegraphics[width=0.34\textwidth,height=6cm]{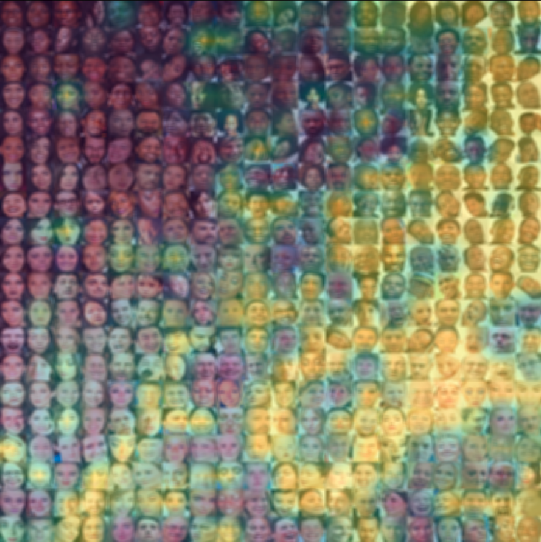}}
\subfloat[ResNet-NTHU-drosy]{\includegraphics[width=0.34\textwidth,height=6cm]{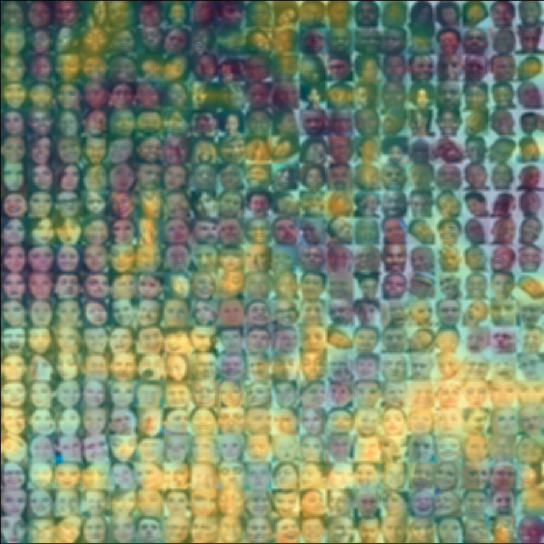}}
\caption{The figure shows images produced using the population bias visualisation technique. All the trained models appear to be failing on the population groups on the upper part of the image. The yellow shaded parts indicate where the model performs well, while failures are indicated by the purple shaded parts, which appear mostly on the upper part of the images. The green shaded parts show that the model is also performing well, but with lower probability (0.50 to 0.65). \label{fig4}}
\end{figure*}
\begin{figure*}
\centering
\subfloat[First visualisation]{\includegraphics[width=0.34\textwidth,height=6cm]{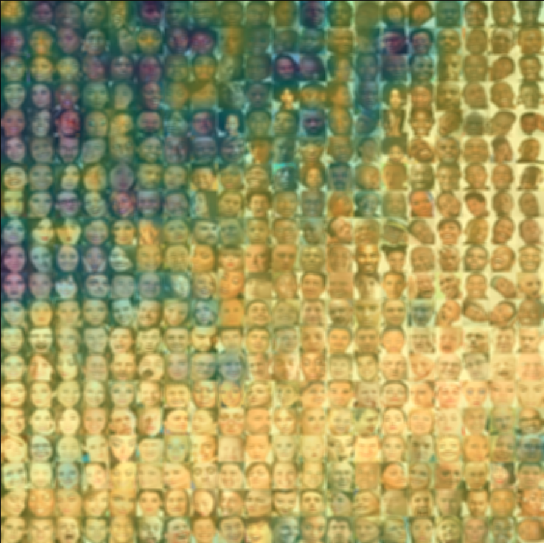}}
\subfloat[Second visualisation]{\includegraphics[width=0.34\textwidth,height=6cm]{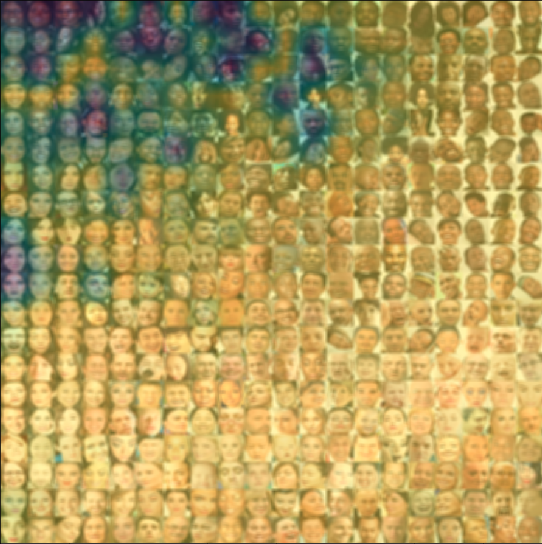}}
\subfloat[Seventh visualisation]{\includegraphics[width=0.34\textwidth,height=6cm]{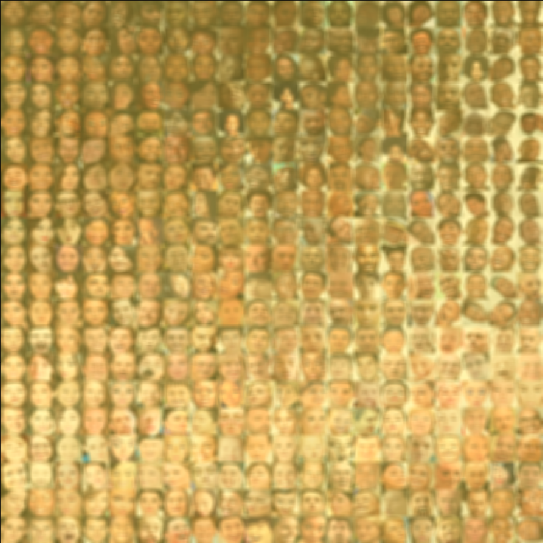}}
\caption{The figure shows the progressive improvement using the framework. As the models were fine-tuned and the learning rate was modified, there was improvement and the models reached more certain classification probabilities (0.70 to 0.99). At the seventh cycle we managed to reach optimal performance on the validation set. \label{fig5}}
\end{figure*}
\begin{figure*}
\centering
\subfloat[First random visualisation]{\includegraphics[width=0.34\textwidth,height=6cm]{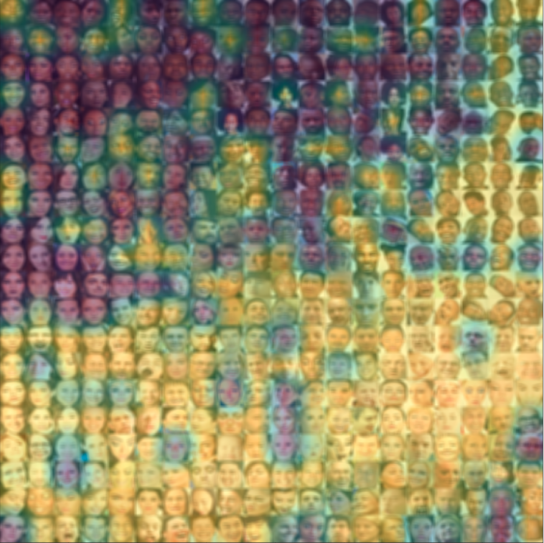}}
\subfloat[Second random visualisation]{\includegraphics[width=0.34\textwidth,height=6cm]{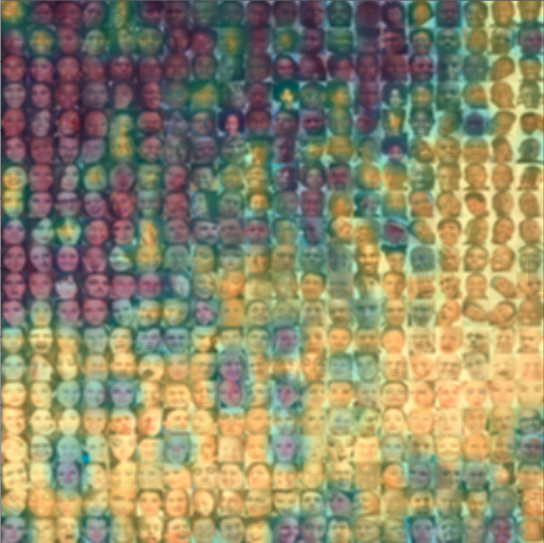}}
\subfloat[Seventh random visualisation]{\includegraphics[width=0.34\textwidth,height=6cm]{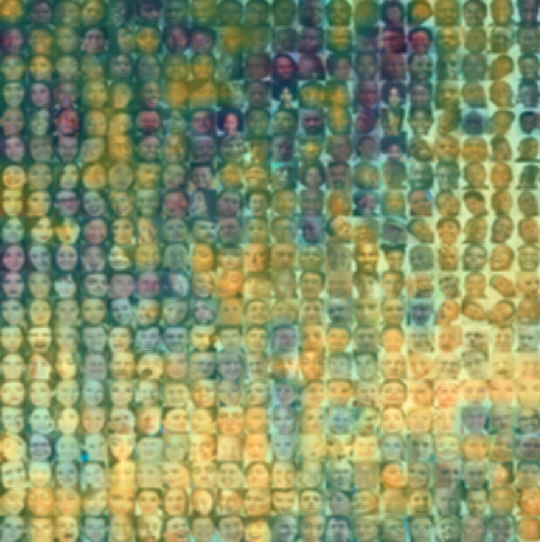}}
\caption{The figure shows the results when the random selection strategy is used without targeted sampling. We randomly selected GAN generated images and this shows the contribution of the image to the performance of the network.  \label{random}}
\end{figure*}

\subsection{Population Bias Visualisation Results}
 
 Population bias visualisation highlights faces where the model fails to generalise. A re-sampling strategy randomly targets highlighted images where the model is failing and uses similar GAN images for model retraining. In these experiments, we compared our framework to models that are trained on CEW, NTHU-drowsy, and DROZY datasets. These models are then tested on the prepared test set. The GAN synthetic data was generated from the validation set which is prepared to represent a wide variety of races. These synthetic generated images cover a wide variety of races and facial attributes.   \par
\begin{table}
\begin{center}
\caption{Model classification accuracy}
\label{table2}
\begin{tabular}{|p{1.0cm}|p{1.0cm}|p{1.5cm}|p{1.5cm}|p{1.5cm}|}
 \hline
 \multicolumn{5}{|l|}{Detection Accuracy} \\
 \hline
  Iteration& Original Data & GAN Augmentation (validation) & GAN Augmentation with Sampling (validation) & GAN Augmentation with Sampling (test) \\
 \hline
  1& 56.40\% & 74.70\% & 87.65\% & 85.28\% \\
 2& 55.70\%&  76.86\%   & 89.87\% & 85.97\%\\
 3& 57.08\%& 77.91\% & 89.43\% & 86.42\% \\
 4& 52.00\% &77.01\% & 90.04\% & 87.02\%\\
5&  59.76\%&  78.80\%   & 93.66 & 88.83\%\\
6&  60.92\%& 80.51\% & 94.5\%4 & 89.05\%\\
7&  60.54\% &80.93\% & 96.75\% & 91.62\% \\
 \hline
\end{tabular}
\end{center}
\end{table}
Fig \ref{random} shows baseline experimental results based where we sampled randomly from the augmentation data. It was noted that when we used all the GAN generated images and did not use targeted sampling, model improvement was limited. Targeted sampling is clearly a more effective strategy.

Fig \ref{fig4} shows the population bias visualisation for the three datasets. It is clear that the models were failing to generalise on darker complexion ethnicity groups, but the proposed framework corrects this. Table \ref{table2} shows the accuracy results from the three experiments which show how accuracy is improved with additional iterations of targeted sampling. The table also contains results from previously unseen test images as the model is updated using the sampling technique, which show improvements in line with the validation set. 

Fig \ref{fig5} highlights the improvement of the driver drowsiness detection performance using the proposed framework. Given enough data, the model was able to generalise well to all population groups in the test set. 

\subsection{Learning rate Results}
Table \ref{learning Rates}. shows the influence of different learning rates on the performance of the model. A learning rate of ${\rm 1e}^{-3}$ was too high and limited model learning. The best performance was observed from ${\rm 1e}^{-4}$ to ${\rm 1e}^{-6}$. The early stopping strategy was also applied to prevent overfitting which can improve the generalisation on the model. At first, the models were trained with the same number of epochs. It was observed that using the same number of epochs on different learning rates showed traits of overfitting. Therefore, early stopping was used when there is no change in the accuracy. 
\begin{table}
\begin{center}
\caption{Learning rate - accuracy tradeoff}
\label{learning Rates}
\begin{tabular}{ |p{2cm}||p{2cm}| }
 \hline
 \multicolumn{2}{|c|}{Learning rates with accuracies} \\
 \hline
 Learning Rate& Accuracies\\
 \hline
 ${\rm 1e}^{-3}$  & 89.70\%   \\
 ${\rm 1e}^{-4}$&   96.30\%    \\
 ${\rm 1e}^{-5}$& 96.91\% \\
 ${\rm 1e}^{-6}$ &98.01\%  \\
 \hline
\end{tabular}
\end{center}
\end{table}

\section{CONCLUSION}\label{CONCLUSION}
In this paper, we introduced a novel framework that can be used to boost the performance of driver drowsiness detection models by reducing bias in the training dataset. Here, a GAN network produces realistic images that are used when retraining a ResNet model on synthetic data generated based on failure cases in validation data.

Faces where the model fails are used to search for similar images in a synthetic dataset produced by the GAN network. These images are used to fine tune a CNN model, and this process is repeated until convergence. A strategy of choosing different learning rates for each iteration helped to prevent the chances of overfitting the model.

Importantly, the proposed approach does not rely on any meta-data or assumptions about the race or ethnicity of individuals in the datasets, which is a commonly used approach to determine algorithmic fairness or bias. Requiring this knowledge is potentially problematic as it tends to rely on subjective and controversial racial classifications. 

This work has shown that bias in datasets can be addressed to some extent using targeted sampling and generative adversarial networks. However, this process still requires that some training data be available for various population groups, and does not eliminate the need for good, representative datasets. Rather, the proposed approach is intended to remedy more subtle bias introduced by imbalanced datasets, where images of a particular group may be more numerous than that of another.

\small{
\bibliographystyle{IEEEtran}
\bibliography{refs.bib}

\begin{thebibliography}{10}
\providecommand{\url}[1]{#1}
\csname url@samestyle\endcsname
\providecommand{\newblock}{\relax}
\providecommand{\bibinfo}[2]{#2}
\providecommand{\BIBentrySTDinterwordspacing}{\spaceskip=0pt\relax}
\providecommand{\BIBentryALTinterwordstretchfactor}{4}
\providecommand{\BIBentryALTinterwordspacing}{\spaceskip=\fontdimen2\font plus
\BIBentryALTinterwordstretchfactor\fontdimen3\font minus
  \fontdimen4\font\relax}
\providecommand{\BIBforeignlanguage}[2]{{%
\expandafter\ifx\csname l@#1\endcsname\relax
\typeout{** WARNING: IEEEtran.bst: No hyphenation pattern has been}%
\typeout{** loaded for the language `#1'. Using the pattern for}%
\typeout{** the default language instead.}%
\else
\language=\csname l@#1\endcsname
\fi
#2}}
\providecommand{\BIBdecl}{\relax}
\BIBdecl

\bibitem{globalreport}
\BIBentryALTinterwordspacing
``{Global status report on road safety 2018}.'' [Online]. Available:
  \url{https://www.who.int/violence_injury_prevention/road_safety_status/2018/English-Summary-GSRRS2018.pdf}
\BIBentrySTDinterwordspacing

\bibitem{driver2019}
\BIBentryALTinterwordspacing
``{Driver fatigue},'' 2019. [Online]. Available:
  \url{http://driverfatigue.co.uk/}
\BIBentrySTDinterwordspacing

\bibitem{WHO2019}
\BIBentryALTinterwordspacing
``{Global status report on road safety 2018},'' 2019. [Online]. Available:
  \url{https://www.who.int/violence_injury_prevention/road_safety_status/2018/en/}
\BIBentrySTDinterwordspacing

\bibitem{Africa2019}
\BIBentryALTinterwordspacing
``{Africa has 2\% of world's cars but 20\% of road deaths'-first safety
  observatory to curb horrendous death toll},'' 2019. [Online]. Available:
  \url{https://www.wheels24.co.za}
\BIBentrySTDinterwordspacing

\bibitem{Knipling1994}
R.~R. Knipling, ``{Vehicle-Based Drowsy Driver Detection : Current Status and
  Future Prospects},'' 1994.

\bibitem{Sahayadhas2012}
A.~Sahayadhas, K.~Sundaraj, and M.~Murugappan, ``{Detecting Driver Drowsiness
  Based on Sensors: A Review},'' pp. 16\,937--16\,953, 2012.

\bibitem{Luthra2015}
A.~Luthra, \emph{{ECG Made Easy}}, 2015, vol.~1.

\bibitem{Drewes2000}
\BIBentryALTinterwordspacing
C.~Drewes, ``{Electromyography: Recording electrical signals from human
  muscle},'' \emph{Tested Studies for Laboratory Teaching. Association for
  {\ldots}}, pp. 248--270, 2000. [Online]. Available:
  \url{http://ableweb.org/volumes/vol-21/12-drewes.pdf}
\BIBentrySTDinterwordspacing

\bibitem{Folane2016}
N.~R. Folane and R.~Autee, ``{EEG Based Brain Controlled Wheelchair for
  Physically Challenged People.}'' \emph{International Journal of Innovative
  Research in Computer and Communication Engineering}, vol.~4, no.~6, pp.
  2257--2263, 2016.

\bibitem{Awais}
M.~Awais, N.~Badruddin, and M.~Drieberg, ``{A Hybrid Approach to Detect Driver
  Drowsiness Utilizing Physiological Signals to Improve System Performance and
  Wearability},'' pp. 1--16.

\bibitem{Man2015}
L.~Man and M.~Hui-ling, ``{A Method of Driver Fatigue Detection based on
  Multi-features},'' vol.~8, no.~10, pp. 107--114, 2015.

\bibitem{Rajput2013}
M.~V. Rajput, ``{Execution Scheme for Driver Drowsiness Detection using Yawning
  Feature},'' vol.~62, no.~6, pp. 6--11, 2013.

\bibitem{JACOBEDENAUROIS2017}
\BIBentryALTinterwordspacing
C.~J. de~Naurois, C.~Bourdin, A.~Stratulat, E.~Diaz, and J.-L. Vercher,
  ``Detection and prediction of driver drowsiness using artificial neural
  network models,'' \emph{Accident Analysis \& Prevention}, 2017. [Online].
  Available:
  \url{http://www.sciencedirect.com/science/article/pii/S0001457517304347}
\BIBentrySTDinterwordspacing

\bibitem{8261140}
M.~{Ngxande}, J.~{Tapamo}, and M.~{Burke}, ``Driver drowsiness detection using
  behavioral measures and machine learning techniques: A review of state-of-art
  techniques,'' in \emph{2017 Pattern Recognition Association of South Africa
  and Robotics and Mechatronics (PRASA-RobMech)}, Nov 2017, pp. 156--161.

\bibitem{8279704}
R.~{Shima}, H.~{Yunan}, O.~{Fukuda}, H.~{Okumura}, K.~{Arai}, and N.~{Bu},
  ``Object classification with deep convolutional neural network using spatial
  information,'' in \emph{2017 International Conference on Intelligent
  Informatics and Biomedical Sciences (ICIIBMS)}, Nov 2017, pp. 135--139.

\bibitem{Jangid2018}
S.~Jangid, ``{Semantic Image Segmentation using Deep Convolutional Neural
  Networks and Super-Pixels},'' vol.~13, no.~20, pp. 14\,657--14\,663, 2018.

\bibitem{7978672}
D.~{Li} and W.~{Chen}, ``Object tracking with convolutional neural networks and
  kernelized correlation filters,'' in \emph{2017 29th Chinese Control And
  Decision Conference (CCDC)}, May 2017, pp. 1039--1044.

\bibitem{detecting-inter}
M.~Ngxande, J.-R. Tapamo, and M.~Burke, ``Detecting inter-sectional accuracy
  differences in driver drowsiness detection algorithms,'' \emph{arXiv preprint
  arXiv:1904.12631}, 2019.

\bibitem{goodfellow2014generative}
I.~Goodfellow, J.~Pouget-Abadie, M.~Mirza, B.~Xu, D.~Warde-Farley, S.~Ozair,
  A.~Courville, and Y.~Bengio, ``Generative adversarial nets,'' in
  \emph{Advances in neural information processing systems}, 2014, pp.
  2672--2680.

\bibitem{IsolaZZE16}
\BIBentryALTinterwordspacing
P.~Isola, J.~Zhu, T.~Zhou, and A.~A. Efros, ``Image-to-image translation with
  conditional adversarial networks,'' \emph{CoRR}, vol. abs/1611.07004, 2016.
  [Online]. Available: \url{http://arxiv.org/abs/1611.07004}
\BIBentrySTDinterwordspacing

\bibitem{1811-07056}
\BIBentryALTinterwordspacing
J.~Ngiam, D.~Peng, V.~Vasudevan, S.~Kornblith, Q.~V. Le, and R.~Pang, ``Domain
  adaptive transfer learning with specialist models,'' \emph{CoRR}, vol.
  abs/1811.07056, 2018. [Online]. Available:
  \url{http://arxiv.org/abs/1811.07056}
\BIBentrySTDinterwordspacing

\bibitem{00219}
\BIBentryALTinterwordspacing
X.~Wang, K.~Yu, S.~Wu, J.~Gu, Y.~Liu, C.~Dong, C.~C. Loy, Y.~Qiao, and X.~Tang,
  ``{ESRGAN:} enhanced super-resolution generative adversarial networks,''
  \emph{CoRR}, vol. abs/1809.00219, 2018. [Online]. Available:
  \url{http://arxiv.org/abs/1809.00219}
\BIBentrySTDinterwordspacing

\bibitem{00676}
\BIBentryALTinterwordspacing
C.~Bodnar, ``Text to image synthesis using generative adversarial networks,''
  \emph{CoRR}, vol. abs/1805.00676, 2018. [Online]. Available:
  \url{http://arxiv.org/abs/1805.00676}
\BIBentrySTDinterwordspacing

\bibitem{Arjovsky2017}
M.~Arjovsky, S.~Chintala, and L.~Bottou, ``{Wasserstein GAN},''
  \emph{Proceedings of the 34th International Conference on Machine Learning
  (ICML)}, vol.~70, pp. 214--223, 2017.

\bibitem{GulrajaniAADC17}
\BIBentryALTinterwordspacing
I.~Gulrajani, F.~Ahmed, M.~Arjovsky, V.~Dumoulin, and A.~C. Courville,
  ``Improved training of wasserstein gans,'' \emph{CoRR}, vol. abs/1704.00028,
  2017. [Online]. Available: \url{http://arxiv.org/abs/1704.00028}
\BIBentrySTDinterwordspacing

\bibitem{RadfordMC15}
\BIBentryALTinterwordspacing
A.~Radford, L.~Metz, and S.~Chintala, ``Unsupervised representation learning
  with deep convolutional generative adversarial networks,'' \emph{CoRR}, vol.
  abs/1511.06434, 2015. [Online]. Available:
  \url{http://arxiv.org/abs/1511.06434}
\BIBentrySTDinterwordspacing

\bibitem{mirza2014conditional}
M.~Mirza and S.~Osindero, ``Conditional generative adversarial nets,''
  \emph{arXiv preprint arXiv:1411.1784}, 2014.

\bibitem{6117593}
J.~{Parris}, M.~{Wilber}, B.~{Heflin}, H.~{Rara}, A.~{El-barkouky}, A.~{Farag},
  J.~{Movellan}, , M.~{Castrilón-Santana}, J.~{Lorenzo-Navarro}, M.~N. {Teli},
  S.~{Marcel}, C.~{Atanasoaei}, and T.~E. {Boult}, ``Face and eye detection on
  hard datasets,'' in \emph{2011 International Joint Conference on Biometrics
  (IJCB)}, Oct 2011, pp. 1--10.

\bibitem{Gupta2019}
R.~Gupta, ``Data augmentation for low resource sentiment analysis using
  generative adversarial networks,'' 2019.

\bibitem{s11291}
\BIBentryALTinterwordspacing
T.~C.~W. Mok and A.~C.~S. Chung, ``Learning data augmentation for brain tumor
  segmentation with coarse-to-fine generative adversarial networks,''
  \emph{CoRR}, vol. abs/1805.11291, 2018. [Online]. Available:
  \url{http://arxiv.org/abs/1805.11291}
\BIBentrySTDinterwordspacing

\bibitem{Wu}
\BIBentryALTinterwordspacing
E.~Wu, K.~Wu, D.~Cox, and W.~Lotter, ``Conditional infilling gans for data
  augmentation in mammogram classification,'' \emph{CoRR}, vol. abs/1807.08093,
  2018. [Online]. Available: \url{http://arxiv.org/abs/1807.08093}
\BIBentrySTDinterwordspacing

\bibitem{Antoniou}
A.~Antoniou, A.~Storkey, and H.~Edwards, ``{Augmenting Image Classifiers using
  Data Augmentation Generative Adversarial Networks},'' \emph{2017}, pp. 1--10.

\bibitem{Shadowen2017}
A.~N. Shadowen, ``{Ethics and Bias in Machine Learning : A Technical Study of
  What Makes Us “ Good ”},'' 2017.

\bibitem{Rhue2019}
L.~Rhue, ``{Emotion-reading tech fails the racial bias test},'' no. January,
  pp. 13--15, 2019.

\bibitem{Rekognition2019}
L.~D. Rekognition, W.~Ai, D.~Raji, and A.~Science, ``{Study takes aim at biased
  AI facial- recognition technology},'' no. February, pp. 1--3, 2019.

\bibitem{buolamwini18a}
\BIBentryALTinterwordspacing
J.~Buolamwini and T.~Gebru, ``Gender shades: Intersectional accuracy
  disparities in commercial gender classification,'' in \emph{Proceedings of
  the 1st Conference on Fairness, Accountability and Transparency}, ser.
  Proceedings of Machine Learning Research, S.~A. Friedler and C.~Wilson, Eds.,
  vol.~81.\hskip 1em plus 0.5em minus 0.4em\relax New York, NY, USA: PMLR,
  23--24 Feb 2018, pp. 77--91. [Online]. Available:
  \url{http://proceedings.mlr.press/v81/buolamwini18a.html}
\BIBentrySTDinterwordspacing

\bibitem{Garvie2016}
C.~Garvie, ``{The perpetual line-up: Unregulated police face recognition in
  america},'' 2016.

\bibitem{DBLP1901-09451}
\BIBentryALTinterwordspacing
M.~De{-}Arteaga, A.~Romanov, H.~M. Wallach, J.~T. Chayes, C.~Borgs,
  A.~Chouldechova, S.~C. Geyik, K.~Kenthapadi, and A.~T. Kalai, ``Bias in bios:
  {A} case study of semantic representation bias in a high-stakes setting,''
  \emph{CoRR}, vol. abs/1901.09451, 2019. [Online]. Available:
  \url{http://arxiv.org/abs/1901.09451}
\BIBentrySTDinterwordspacing

\bibitem{Benthall3287575}
\BIBentryALTinterwordspacing
S.~Benthall and B.~D. Haynes, ``Racial categories in machine learning,'' in
  \emph{Proceedings of the Conference on Fairness, Accountability, and
  Transparency}, ser. FAT* '19.\hskip 1em plus 0.5em minus 0.4em\relax New
  York, NY, USA: ACM, 2019, pp. 289--298. [Online]. Available:
  \url{http://doi.acm.org/10.1145/3287560.3287575}
\BIBentrySTDinterwordspacing

\bibitem{Abiteboul:2017}
\BIBentryALTinterwordspacing
S.~Abiteboul, ``Issues in ethical data management,'' in \emph{Proceedings of
  the 19th International Symposium on Principles and Practice of Declarative
  Programming}, ser. PPDP '17.\hskip 1em plus 0.5em minus 0.4em\relax New York,
  NY, USA: ACM, 2017, pp. 1--1. [Online]. Available:
  \url{http://doi.acm.org/10.1145/3131851.3131854}
\BIBentrySTDinterwordspacing

\bibitem{star09020}
\BIBentryALTinterwordspacing
Y.~Choi, M.~Choi, M.~Kim, J.~Ha, S.~Kim, and J.~Choo, ``Stargan: Unified
  generative adversarial networks for multi-domain image-to-image
  translation,'' \emph{CoRR}, vol. abs/1711.09020, 2017. [Online]. Available:
  \url{http://arxiv.org/abs/1711.09020}
\BIBentrySTDinterwordspacing

\bibitem{DepthwiseGANs}
M.~{Ngxande}, J.~{Tapamo}, and M.~{Burke}, ``Depthwisegans: Fast training
  generative adversarial networks for realistic image synthesis,'' in
  \emph{2019 Southern African Universities Power Engineering
  Conference/Robotics and Mechatronics/Pattern Recognition Association of South
  Africa (SAUPEC/RobMech/PRASA)}, Jan 2019, pp. 111--116.

\bibitem{ulyanov1607instance}
D.~Ulyanov, A.~Vedaldi, and V.~Lempitsky, ``Instance normalization: The missing
  ingredient for fast stylization. arxiv 2016,'' \emph{arXiv preprint
  arXiv:1607.08022}.

\bibitem{Li2016}
C.~Li and M.~Wand, ``{Precomputed Real-Time Texture Synthesis with Markovian
  Generative Adversarial Networks},'' pp. 1--17, 2016.

\bibitem{2015arXiv151203385H}
K.~{He}, X.~{Zhang}, S.~{Ren}, and J.~{Sun}, ``{Deep Residual Learning for
  Image Recognition},'' \emph{arXiv e-prints}, p. arXiv:1512.03385, Dec 2015.

\bibitem{Smith2002}
L.~I. Smith, ``{A tutorial on Principal Components Analysis},''
  \emph{Statistics}, vol.~51, p.~52, 2002.

\bibitem{Stewart}
\BIBentryALTinterwordspacing
G.~Stewart, ``On the early history of the singular value decomposition,''
  \emph{SIAM Review}, vol.~35, no.~4, pp. 551--566, 1993. [Online]. Available:
  \url{https://doi.org/10.1137/1035134}
\BIBentrySTDinterwordspacing

\bibitem{2014arXiv1412.6980K}
D.~P. {Kingma} and J.~{Ba}, ``{Adam: A Method for Stochastic Optimization},''
  \emph{arXiv e-prints}, p. arXiv:1412.6980, Dec 2014.

\bibitem{NTHU}
\BIBentryALTinterwordspacing
``{NTHU CVlab - Driver Drowsiness Detection Dataset},'' 2016. [Online].
  Available: \url{http://cv.cs.nthu.edu.tw/php/callforpaper/datasets/DDD/}
\BIBentrySTDinterwordspacing

\bibitem{massoz2016ulg}
Q.~Massoz, T.~Langohr, C.~Fran{\c{c}}ois, and J.~G. Verly, ``The ulg
  multimodality drowsiness database (called drozy) and examples of use,'' in
  \emph{Applications of Computer Vision (WACV), 2016 IEEE Winter Conference
  on}.\hskip 1em plus 0.5em minus 0.4em\relax IEEE, 2016, pp. 1--7.

\bibitem{ClosedEyes}
\BIBentryALTinterwordspacing
``{The Closed Eyes in the Wild (CEW) dataset}.'' [Online]. Available:
  \url{http://parnec.nuaa.edu.cn/xtan/data/ClosedEyeDatabases.html}
\BIBentrySTDinterwordspacing

\end{thebibliography}
}

\end{document}